\documentclass[11pt]{article}
\usepackage{fullpage}
\usepackage{hyperref}
\usepackage[numbers]{natbib}
\usepackage{graphicx}
\usepackage{caption}
\usepackage{subcaption}
\usepackage{amsmath}
\usepackage{amsthm}
\usepackage[ruled, linesnumbered]{algorithm2e}
\usepackage{amsfonts}
\usepackage{bbm}
\usepackage{amssymb}
\usepackage[utf8]{inputenc}
\makeatletter
\newtheorem*{rep@theorem}{\rep@title}
\newcommand{\newreptheorem}[2]{%
\newenvironment{rep#1}[1]{%
 \def\rep@title{#2 \ref{##1}}%
 \begin{rep@theorem}}%
 {\end{rep@theorem}}}
\makeatother
\newtheorem{thm}{Theorem}
\newreptheorem{thm}{Theorem}
\newtheorem{lem}[thm]{Lemma}
\newreptheorem{lem}{Lemma}
\newtheorem{prop}[thm]{Proposition}
\newreptheorem{prop}{Proposition}

\newreptheorem{cor}{Corollary}
\usepackage{xcolor}
\usepackage{graphicx}

\DeclareMathOperator*{\argmin}{arg\,min}

\newtheorem{defn}[thm]{Definition}

\usepackage{soul}

\def\c{\mathcal L}
\def\d{\Lambda}
\def\ind{\mathbbm{1}}
\def\oc{Online\_Cluster}
\def\mem{\mathcal M}
\def\E{\mathbb{E}}
\def\R{\mathbb{R}}

\title{Online k-means Clustering on Arbitrary Data Streams}

\author{Robi Bhattacharjee \\  rcbhatta@eng.ucsd.edu
   \and Sanjoy Dasgupta \\  dasgupta@eng.ucsd.edu
   \and Jacob John Imola \\  jimola@eng.ucsd.edu
   \and Michal Moshkovitz \\ mmoshkovitz@eng.ucsd.edu }

\date{\today}

\begin{document}

\maketitle

\begin{abstract}
We consider $k$-means clustering in an online setting where each new data point is assigned to its closest cluster center and incurs a loss equal to the squared distance to that center, after which the algorithm is allowed to update its centers. The goal over a data stream $X$ is to achieve a total loss that is not too much larger than $L(X, OPT_k)$, the best possible loss using $k$ fixed centers in hindsight.

We start by introducing a data parameter, $\Lambda(X)$, such that for any algorithm that maintains $O(k \, \text{poly}(\log n))$ centers after seeing $n$ points, there exists a data stream $X$ for which a loss of $\Omega(\d(X))$ is inevitable.
We then give a randomized algorithm that achieves total loss $O(\Lambda(X) + L(X, OPT_k))$. Our algorithm uses $O(k \, \text{poly}(\log n))$ memory and maintains $O(k \, \text{poly}(\log n))$ cluster centers. It has a running time of $O(k \, \text{poly}(\log n))$ and is the first algorithm to achieve polynomial space and time complexity in this setting. It also is the first to have provable guarantees without making any assumptions on the input data.
\end{abstract}
\newpage 
\section{Introduction}

The \emph{online learning} framework~\citep{L87}, first introduced in the context of classification, is a model that does away with the benign (i.i.d.) statistical assumptions that underlie much of learning theory, and instead deals with data that is arbitrary and possibly adversarial, and that arrives one point at a time, indefinitely.

Here we consider \emph{clustering} in an online setting. At every time $t$, the Learner announces a clustering; then, Nature provides the next data point $x_t$; finally, the Learner incurs a loss depending on how well its clustering captures $x_t$. There are no assumptions on the data.

Specifically, we look at online realizations of \emph{k-means clustering}. For any data stream $X = \{x_1, \ldots, \} \subset \R^d$, the $k$-means cost of a set of \emph{centers} $S \subset \R^d$ is $\c(X, S) = \sum_{x \in X} d(x,S)^2$
where we define $d(x,S)$ to be the Euclidean distance from $x$ to its nearest neighbor in $S$.

In the (batch) $k$-means problem, the input is a full data stream $X$, and the goal is to find a set of centers whose cost is close to that of the optimal $k$ centers, denoted by
$$\c_k(X) = \inf_{|S|=k} \c(X,S).$$
Finding the optimal clustering is NP-hard~\citep{aloise2009np,dasgupta2008hardness}, but a variety of constant-factor approximation algorithms are known~\citep{kanungo2004local,ANSW20}.

Batch $k$-means is the canonical method for \emph{vector quantization}, in which training data $X$ is clustered to obtain a set of $k$ \emph{codewords} $S$, and any subsequent data point $x$ is quantized by replacing it by its nearest codeword $s \in S$, at a quantization cost of $d(x,s)^2$. It is well-known that if all points (past and future) come from some fixed distribution, and if $|X|$ is large enough, then good codewords for $X$ are also good codewords for this underlying distribution~\cite{P81,P82}. We are interested in the more challenging setting of \emph{lifelong learning}, where the data distribution can change and thus codewords need to be updated from time to time.

We study the \emph{online $k$-means setting}, in which the Learner maintains a set of centers $S_t$ that it updates as it sees more data. At each time $t$,
\begin{itemize}
\item Nature provides a data point $x_t$
\item For $t \geq 2$, Learner incurs loss $d(x_t, S_{t-1})^2$
\item Learner announces an updated set of centers $S_t$
\end{itemize}
The total loss incurred by the learner up to time $t$ is then compared to the loss of the best $k$-means solution in hindsight; that is, $\c_k(\{x_1, \ldots, x_t\})$. A crucial difficulty is that the data $x_t$ are arbitrary: Nature can choose a stream $X$ with full knowledge of the Learner's algorithm.

There has been a large body of work on the different, but related, \emph{streaming} $k$-means problem. In this setting, there is a finite data stream whose size $n$ is typically known in advance. These data are revealed one point at a time and the learner updates its model after seeing each successive new data item. The key requirements are that the learner not use too much memory and that the individual updates be efficient. Once the entire data stream has been processed in this way, the cost of learner's final model is compared to that of the optimal $k$-means clustering.

In contrast, for the \emph{online} $k$-means problem, losses accumulate along the way, and crucially, the loss of $x_t$ depends on a clustering that was produced \emph{before} $x_t$ had been seen. The only work we are aware of in this framework is that of \cite{CGKR21}. They show how the classical multiplicative weights strategy~\citep{LW94} can be used in this setting, with each candidate clustering being an \emph{expert}. This space of experts is continuous, but it can be discretized, and the authors show how to do this in a way that leads to a strong performance guarantee: the learner outputs exactly $k$ centers at each time step and the total loss it accumulates at each time $n$ is at most $(1+\epsilon) \c_k(\{x_1, \ldots, x_n\})$. The downside, however, is that the algorithm requires resources (space and time) that are exponential in $k$ and $d$, making it impractical in many settings.

In this paper, we are interested in developing \emph{efficient} algorithms for online $k$-means clustering. Our solution strategy does not use multiplicative weights. Instead we rely upon three key ideas.

First, the adversarial nature of the data means that the \emph{scale} of the clustering problem can increase dramatically from time to time, for instance if the latest point $x_t$ is much farther away from the rest of the data than the typical previous interpoint distance. Between such scale-changes, however, it turns out that algorithmic ideas from the streaming $k$-means literature are applicable.

Second, we make use of the availability of good algorithms for the \emph{streaming $k$-center} problem. For a data stream $X$ and a set of centers $S$, the $k$-center cost of $S$ is the maximum distance from a point in $X$ to its closest center in $S$.
The algorithm of \cite{Charikar97} for streaming $k$-center takes one data point $x_t$ at a time and updates its set of $k$ centers in $O(kd)$ time. Its total space requirement is just $O(kd)$. And at any time $n$, this set of $k$ centers has cost at most eight times that of the best $k$-center solution for $X_n = \{x_1, \ldots, x_n\}$. This does not give us a solution for the $k$-means problem, since the $k$-means and $k$-center cost functions can differ by a multiplicative factor of $\Omega(n)$ for $n$ data points. However, the $k$-center cost is useful for gauging when there has been a dramatic \emph{scale-change}. We run the streaming $k$-center algorithm in the background, and whenever the $k$-center cost increases sharply, we think of a new \emph{scale} as having begun.

Third, it is necessary to periodically throw away centers when we have accumulated too many of them. This is tricky because we must always ensure that the data points $x_t$ close to those centers are still adequately covered. We introduce a novel way of handling this: we throw away all centers from before the previous scale began, and replace them by the $k$-center solution so far. The nature of scale-change means that quantization error can still be controlled.

Our algorithm is shown in Algorithm~\ref{alg:main}, and its performance is governed by our main result.

\begin{thm}[\textbf{Upper Bound}]\label{thm:main}
Let $X$ be an arbitrary data sequence, $k$ be a positive integer, and $\delta$ satisfy $0 < \delta < 1$.
Suppose we run $\oc(X, k, \delta)$ (Algorithm~\ref{alg:main}). Let $S_t$ denote the centers outputted at time $t$ and $\mem_t$ denote the
total amount of memory used at the end of time $t$. Then with probability at least $1 - \delta$ over the randomness of $\oc$, for all integers $n \geq 2$, the following hold:
\begin{enumerate}
	\item (Approximation Factor) $\sum_{t = 2}^n d(x_t, S_{t-1})^2 = O(\c_k(X_n) + \d(X_n))$.
	\item (Center Complexity) $|S_n| = O(k\log^6 \frac{n}{\delta})$.
	\item (Memory and Time Complexity) Each step uses $O(kd \log^6 \frac{n}{\delta})$ time and memory.
\end{enumerate}
\end{thm}

Here, $X_n$ is a shorthand for the sequence $x_1, \ldots, x_n$, $\c_k(X_n)$ is the optimal $k$-means cost in hindsight, and the final term is
$$ \Lambda(X_n) = \sum_i d(x_i, X_{i-1})^2.$$
The last term, $\Lambda(X_n)$, can be seen as the loss we would incur, in the online setting, if we were allowed to store \emph{all} points seen so far. We complement this with a lower bound demonstrating a broad class of data sequences for which at least $\Omega(\d(X_n))$ loss must be paid over the first $n$ points.

\begin{thm}[\textbf{Lower Bound}]\label{thm:lower}
Let $X$ be any data sequence that contains infinitely many distinct points. Let $A$ be an online clustering algorithm such that its output satisfies $|S_n| < n$ for all $n$ and for all input sequences. Then there exists a sequence $\tilde{X} = \tilde{x}_1, \tilde{x_2}, \dots$ such that the following conditions hold.
\begin{enumerate}
	\item $\tilde{X}$ is drawn from the closure of $X$, (i.e. $X$ and its limit points). Thus all points in $\tilde{X}$ are arbitrarily close to points in $X$.
	\item For all $n \geq 2$, the expected loss over $A$ satisfies $\mathbb{E}_A\left[ \sum_{s = 2}^n d(\tilde{x}_s, S_{s-1})^2\right] \geq \Omega(\d(\tilde{X}_n))$.
\end{enumerate}
\end{thm}

Our lower bound does not construct a fixed sequence for which \emph{all} algorithms incur a large loss, for the simple reason that an algorithm may memorize an arbitrary number of the points in the sequence. Thus the sequences achieving the lower bound do depend on the algorithm, but they are not pathological in the sense that they can be constructed from any sequence $X$ with infinitely many distinct points.
\section{Related Work}
In the offline (batch) $k$-means setting, all points are given simultaneously, and the goal is to find a small subset of centers with a small approximation compared to the optimal $k$-means clustering. Efficient algorithms returning $poly(k)$ centers with a constant approximation are known \citep{kanungo2004local,aggarwal2009adaptive,ANSW20}. On the negative side, it is NP-hard to return the optimal $k$ centers or even approximate it up to a small constant \citep{aloise2009np,dasgupta2008hardness}.

In the online setting, points arrive one after another and not simultaneously. After each point, the algorithm decides whether to take this point as a center. In~\cite{CGKR21} algorithms for a similar setting as ours were proposed. The algorithms are inefficient and run in exponential time in $k$ and the dimension of the data. On the other hand, our algorithm runs in polynomial time.

Recently a popular variant of the online setting was explored, the no-substitution setting. In this setting, decisions are irrevocable. Additionally, the cost is measured only at the end \citep{hess2020sequential,moshkovitz2021unexpected,bhattacharjee2021no,hess2021constant}.  In \cite{liberty2016algorithm} an algorithm utilizing ideas from \cite{meyerson01} was introduced. Unfortunately, the number of centers inherently depends on the aspect ratio, which can be enormous.  In this paper, the cost is incurred immediately after each point arrives.
The number of centers our algorithm uses is only $k\log^{O(1)}(n)$. No assumptions on the input data are needed.

A closely related setting is the streaming model \citep{aggarwal2007data,guha03,braverman11,shindler11,ailon09}. As in the online setting, points arrive one after another, and the goal is the utilize a small memory while ensuring that the cost at the end is small.
Several passes over the data are allowed. In this paper, only one pass on the data is allowed, and more importantly, the cost is incurred immediately and not at the end of the stream.
\section{Preliminaries}

\paragraph{Notation} For the rest of this paper, we prefer referring to data streams as data sequences to emphasize their possibly infinite size. For a data sequence $X \subseteq \R^d$, the $k$-means cost of a set of centers
$S \subseteq \R^d$ is given by $$\c(X, S) = \sum_{x \in X} \min_{s \in S} d(x, s)^2,$$
where $d$ denotes the Euclidean distance. Additionally, we let $\c_k(X) = \min_{|S| = k} \c(X, S)$ denote the optimal
$k$-means cost of clustering data sequence $X$.

For a positive integer $t$, we write $X_t = \{x_1, \ldots, x_t\}$
to denote the first $t$ elements of $X$. We also let $\c(X)$ and $\c(X, x)$
denote $\c_1(X)$ and $\c(X, \{x\})$ respectively.

\paragraph{Setting} Our input is an infinite sequence $X = \{x_1, \ldots\} \subset \R^d$ which is given to the algorithm one point at a time. At each time $t$, the algorithm observes a new point $x_t$ and then outputs a set of cluster centers, $S_t$. The algorithm incurs loss at $x_t$ based on how well the \textit{previous} clustering, $S_{t-1}$, captures $x_t$.  Thus, the total loss of the algorithm is $$\sum_{t=2}^n d(x_t, S_{t-1})^2.$$ Here, the index begins at $t=2$ because the algorithm is allowed to see the initial point, $x_1$ without incurring a loss.

There are no restrictions on the sequence $X$. It can even be chosen by an adversary that has knowledge of our algorithm ahead of time. The only restriction is that $X$ cannot be changed \textit{after} observing the output of our algorithm, $S_t$ -- the adversary (or nature) must decide on $X$ before the algorithm starts running.
\section{The Lower Bound Parameter, $\Lambda$}

Typically, the goal of most online or streaming clustering problems is to achieve a loss at time $n$ on the order of $O(\c_k(X_n))$, where $k$ is a pre-specified parameter denoting the optimal number of centers. However, in this setting, this is \textit{not} always possible.

Consider the data sequence $\{1, \alpha, \alpha^2, \alpha^3, \dots \} \subset \R$ for some constant $\alpha > 1$. For $\alpha$ sufficiently large, the point $\alpha^n$ will be extremely far away from all the points preceding it, and consequently will be very likely to incur a large loss, $d(x_n, S_{n-1})^2$. The only way $d(x_n, S_{n-1})^2$ will be small is if the algorithm is somehow able to ``guess" its location. By contrast, the optimal $k$-means clustering of $\{1, \dots, \alpha^n\}$ will include some cluster center that is reasonably close to $\alpha^n$, and as a result will incur a significantly smaller loss.

To generalize this example to arbitrary data sequences, the key insight is that the distance from a given point to the set of points preceding it serves as a baseline for its incurred loss unless the algorithm makes a lucky guess. This leads us to our lower bound parameter.

\begin{defn}[\textbf{Lower Bound Parameter}]\label{defn:lower}
For an ordered sequence of points $X_n = \{x_1, x_2, \dots, x_n\}$, define $\d(X_n) = \sum_{t = 2}^n d(x_t, X_{t-1})^2$.
\end{defn}


The quantity $\Lambda(X_n)$ can be interpreted as the loss incurred by an online algorithm whose cluster centers at any time $t$ consist of \emph{all} points seen so far. One can view this as the best possible algorithm that makes no guesses about locations of future points.

We now formalize the intuition that $\d(X_n)$ is a lower bound on the online loss at time $n$. With no assumptions, the algorithm may make an arbitrary number of guesses about the data sequence, or even memorize $X_n$, and defeat the lower bound of $\d(X_n)$. To control this behavior, we make the mild assumption that at time $n$, the number of centers outputted by the algorithm is at most a fixed integer $b_n$.

The most standard way to prove a lower bound would be to show that for any algorithm $A$, there exists a data sequence $X$ for which $A$ pays loss bounded by $\Omega(\d(X_n))$ at time $n$. However, the lower bound would then only be tight for pathological choices of $X_n$ (such as $\{1, \alpha, \alpha^2, \dots\}$). Instead, we show a stronger result--- that for any algorithm $A$ and data sequence $X$, there exists a data sequence $\tilde{X}$ that can be constructed from $X$ for which $A$ pays loss at least $\Omega(\d(\tilde{X}_n))$. The strength of this stricter approach is that even for extremely limited sets of data sequences (say sets where ``pathological" examples are excluded), our lower bound $\d(X_n)$ maintains relevance. Our lower bound appears in Theorem~\ref{thm:lower}.

\paragraph{Proof idea of Theorem \ref{thm:lower}}

We now summarize the main ideas of the proof of Theorem \ref{thm:lower}. The full proof is deferred to Appendix  \ref{app:low_bound}.

The key idea is to first consider the case in which the input sequence $X$ satisfies some additional structure that allows us to cleanly construct sub-sequences, $\tilde{X}$, with the desired property. We call such inputs \textit{nice} sequences.

\begin{defn}\label{defn:nice_sequence}
A sequence $X$ is \textbf{nice} if it consists of distinct points such that for all $1 < i < j$, $d(x_i, x_j) > \frac{1}{2}d(x_i, x_1)$ and  $d(x_i, x_j) > \frac{1}{2}d(x_j, x_1)$.
\end{defn}

Nice sequences have the property that all of their points are relatively well ``spread out" in comparison to their distance to the first point. Thus, in order for an algorithm to achieve a \textit{better} loss than $\d(X_n)$, the loss incurred by an algorithm that makes no guesses about future points, it must have guessed the value of $x_n$. This is the main idea behind how we construct $\tilde{X}$ from a nice sequence: the adversary randomly chooses its next points from a large set (larger than $b_n$) of points which means that on average, any algorithm is going to fail at guessing. We give a full proof of this case in Lemma \ref{lem:lower_help} in Appendix \ref{app:low_bound}.

Next, to prove the general version of Theorem \ref{thm:lower}, it suffices to reduce a general sequence $X$ to a nice sequence. In particular, we must show that any sequence $X$ has a nice subsequence (with possible rearrangements to the order of the subsequence). To do this, we appeal to the fact that $X$ is a subset of $\R^d$ by using casework on whether $X$ is bounded, and thus contained in a compact set, or unbounded. Both cases follow from a similar type of argument, the only difference is that for bounded sequences, our nice sub-sequence $\tilde{X}$ converges while in the unbounded case it diverges.
\section{Algorithm Description}

Having shown a lower bound of $\Lambda(X_n)$, in this section we present an algorithm with an online loss that satisfies $\sum_{t=2}^n d(x_t, S_{t-1})^2 = O\left(\d(X_n) + \c_k(X_n)\right)$ for all times $n \geq 2$ while outputting $O(k\, \text{poly}\log n)$ centers. Recall that the $\d(X_n)$ term is the lower bound that any algorithm must incur, even an algorithm with perfect memory that takes all previous points as centers. On the other hand, the $\c_k(X_n)$ term is the error of an algorithm that selects the best set of $k$ fixed centers in hindsight. Together, these terms represent the constraints that our algorithm is not allowed to remember every point and must not output too many centers.

\subsection{Motivation: A Naive Application of Streaming Algorithms Does Not Work}
A natural starting point is to consider the vast literature of streaming $k$-means algorithms. Recall that streaming algorithms maintain an output $S_t$ for which the loss $\c_k(X_t, S_t)$ is small (typically $O(\c_k(X_t)$). We might conjecture that such algorithms have an online loss bounded by $O(\d(X_n) + \c_k(X_n))$ for the following reason. When $x_t$ is far from previously encountered points, the incurred loss $d(x_t, S_{t-1})^2$ can be absorbed by $\d(X_n)$, and when $x_t$ is near previous data, the maintained set of centers $S_{t-1}$ serves as a good representation of all data including $x_t$. Unfortunately, the following example illustrates that that this conjecture is not true for all streaming algorithms, highlighting the need for a more sophisticated approach.

Let $X$ be a sequence that cycles through a set of $k+1$ points with pairwise distances all equal to $1$ embedded in $\R^k$. Let $A$ be the algorithm that always outputs the last $k$ points that it has encountered, that is $S_t = \{x_t, x_{t-1}, \dots, x_{t-k+1}\}$. $A$ achieves a good streaming loss over $X$ --- at the end of any time $n$, $A$ clearly outputs a $2$-approximation to the optimal $k$-means loss, and consequently achieves a low streaming loss.

Conversely, $A$ does poorly on $X$ in the online setting. On each subsequent point, $A$ pays a loss of exactly $1$, meaning that our online loss at time $n$ is $\Omega(n)$. By contrast, $\d(X_n) = k$, as we only have $k+1$ distinct points, and $\c_k(X_n) \leq \frac{n}{2(k+1)}$. It follows that $A$ pays online loss that is a factor of $\Omega(k)$ larger than the combination of $\d(X_n)$ and $\c_k(X_n)$.

This example highlights that more structure is needed on the set of centers $S_t$ beyond simply having a low streaming loss. In the example above, at every single time, $A$ removes from $S$ precisely the new point $x_t$ and incurs loss.
The approach our algorithm takes is to rarely remove centers from $S$ so that the performance on previously seen data will not degrade over time. However, we must carefully decide when and how to remove centers so that $S$ does not grow too large and still maintains good performance on previous points. In the next section, we will show how our algorithm resolves this central issue.

\subsection{Our Algorithm}

\begin{algorithm}[h] 
\caption{The main algorithm, $\oc(X, k, \delta)$.}\label{alg:main}
\DontPrintSemicolon
$S_k \leftarrow \{x_1, \dots, x_k\}$ \hfill \texttt{Initial set of centers} \label{line:init}

$R_k,F \leftarrow 0$\;

$\tau_1 \gets k+1, i \gets 2$\;

$(Z_k, w_k) \gets {online\_k\_centers\_update}(X_k)$ \label{line:init2}\;

\For{$t = k+1, k+2, \dots$}{
$(Z_t, w_t) \gets {online\_k\_centers\_update}(x_t)$\hfill \texttt{Scale approximation}\label{line:k_centers}

$time(z) = t$ for $z \in Z_t$\;

\uIf(\hfill\texttt{Scale change detected}){$w_t > 16w_{\tau_{i-1}}\sqrt{t}$}{\label{line:begin_if}

$\tau_{i} \gets t$\;

$S_{t} \gets S_{t-1} \setminus \{x \in S_{t-1} : time(x) \leq \tau_{i-1}\}$
\hfill \texttt{Remove old centers from $S$}\label{line:make_deletion}

$S_{t} \gets S_t \cup Z_{\tau_{i-1}} $
\hfill \texttt{Replace with $k$-centers} \label{line:add_k_centers}

$R_t \leftarrow \frac{w_t^2}{128k}$, $F \leftarrow 0$\; \label{line:updating_old}

$i \gets i+1$\;\label{line:end_if}

}\uElse{$S_t \gets S_{t-1}$, $R_t \gets R_{t-1}$\;
}
With probability $\min\left\{1, \frac{d(x_t, S_t)^2\log^4 \frac{2t}{\delta}}{R_t}\right\}$, $S_t = S_t \cup \{x_t\}$
\hfill \texttt{Center selection} \label{line:adding_point}

$F \leftarrow F + \mathbf{1}_{x_t\text{ is selected}}$\;\label{line:increment_f}

\uIf{$F > 25k\log^5 \frac{2t}{\delta}$}{
$R_t \leftarrow 2R_t$, $F \leftarrow 0$\; \label{line:double_r}
}
}
\end{algorithm}

We now present our algorithm, \oc{}, that with probability $1-\delta$ achieves online loss $\sum_{t=2}^n d(x_t, S_{t-1})^2$ bounded by a combination of $\Lambda(X_n)$ and $\c_k(X_n)$ for all times $n \geq 2$. The pseudocode for the algorithm appears in Algorithm~\ref{alg:main}. On a high level, the algorithm adds centers to $S$ probabilistically and removes older centers when it detects that the scale of the data changes. Specifically, this process involves three ideas: scale approximation, center deletion, and center selection.  We now outline these ideas.

\paragraph{Scale Approximation:}
\oc{} approximates the scale of the data, or how spread out the data is, by running a $k$-centers algorithm in the background at all times. Recall that for a data sequence $X$ and a set of centers $S$, the $k$-center cost of $S$ is $\max_{x \in X}d(x, S)$. The algorithm of \cite{Charikar97} is an online $k$-center algorithm that at any time $n$ outputs a set of $k$ centers $Z_n = \{z_n^1, z_n^2, \dots, z_n^k\}$ with $k$-centers loss at most $8$ times the optimal loss. Furthermore, this algorithm enjoys space and time complexity $O(k)$. While it may seem odd to use a streaming $k$-centers algorithm as opposed to $k$-means, we use $k$-centers because unconditional approximation algorithms exist. We will let $w_n$ denote loss outputted by \cite{Charikar97}'s algorithm when applied to $x_1, \dots, x_n$. 

Scale approximation using $k$-centers is useful for two reasons. First, it enables us to approximate the $k$-means cluster cost $\c_k(X_n)$ at time $n$ up to a factor of $O(n)$. In particular, the $k$-centers clustering ouputed by \cite{Charikar97}'s algorithm can be simply applied as a $k$-means clustering, with the $k$-means loss being bounded by noting that each point incurs loss at most $w_n^2$. Formally, we have the following:

\begin{prop}\label{prop:center_vs_mean}
	For all data sequences $X$ and all $n$, $\frac{w_n^2}{128} \leq \c_k(X_n) \leq nw_n^2$.
	\end{prop}

	\begin{proof}
	The subroutine $online\_k\_centers$ maintains a $k$-centers clustering with cost $w_n$ (at time $n$) that is an $8$-approximation to the optimal $k$-centers cost.

	By directly using the given $k$-centers clustering with cost $w_n$ as a $k$-means clustering, we get an upper bound of $nw_n^2$. Because the lower bound of the $k$-centers cost is $\frac{w_n}{8}$, there must exist two points in any set of $k$ clusters that are clustered together with distance at least $\frac{w_n}{8}$. By the triangle inequality, under any cluster center these two points will incur cost at least $2(\frac{w_n}{16})^2 = \frac{w_n^2}{128}$, which finishes the proof.
	\end{proof}
We will see in the later that despite $w_n^2$ being a loose approximation for $\c_k(X_n)$ (with a gap of up to $O(n)$), it can nevertheless be used to set the center selection rate and prevent too many centers from being selected.

The second reason why scale approximation is useful is that it enables the algorithm to remove selected centers from smaller scales, which brings us to our next key idea.

\paragraph{Removing centers:} \oc{} uses scale \emph{increases} to decide when to remove centers it has previously selected. The key insight is that when the scale, tracked by $w_n$, drastically increases, we have that all previous points can be clustered in ``relatively small" clusters. Because of this, clustering the previous points using their \textit{$k$-centers approximation} provides a sufficient summary. Although the $k$-centers clustering can incur $k$-means loss up to $nw_{n-1}$,  the nature of the scale increase implies that even this total cost is still small compared to the $k$-means cost at time $n$. We will refer to times during which these large scale increases occur as \textit{scale changes}, and denote them as $\tau_1, \tau_2, \dots.$ They have the following formal definition.

\begin{defn}[\textbf{scale changes}]\label{def2:deletion_times}
	Let $\tau_1 = k+1$, and let $\tau_i = \min \{t: t > \tau_{i-1}, w_t > 16\sqrt{t}w_{\tau_{i-1}}\}$. If no such $\tau_i$ exists, then we set $\tau_i = \infty$ and terminate the sequence. Each $\tau_i$ is referred to as a \textbf{scale change}.
\end{defn}

When a scale change $\tau_{i}$ is detected, the algorithm is able to replace all selected centers $x$ that were streamed as inputs before $\tau_{i-1}$ with the set of $k$ centers from that time, denoted by $Z_{\tau_{i-1}}$. To this end, we implicitly assume that the algorithm timestamps each point it selects as a center; this costs a trivial amount of additional time and memory. After removing the desired centers, the only points that will remain in memory are those centers taken after time $\tau_{i-1}$ and $Z_{\tau_{i-1}}$. This prevents the number of selected centers from accumulating over increasing scales.

\paragraph{Center Selection:} The algorithm selects centers using existing ideas from the streaming setting with one important change. Our method resembles that of \cite{liberty2016algorithm} which is the following: for each subsequent point, select it with probability $O(\frac{d(x_t, S_{t-1})^2}{R_t})$, where $R_t$ is a dynamically adjusted parameter governing the rate of selections. Use a counter $F$ to keep track of the number of centers selected since $R_t$ was previously changed. When $F \geq O(k\log t)$, this indicates that the value of $R_t$ is too small, and consequently double $R_t$ so as to discourage further center selection. When $R_t$ reaches a value of $O(\frac{\c_k(X_t)}{k})$, we can prove the desired center complexity and approximation ratio bounds.

Unfortunately, the previous method is incapable of dealing with data that exhibits many scale changes. Consider again the data sequence $\{1, \alpha, \alpha^2, \dots \}$. In this sequence, naively applying the above center selection criteria would result in every point being selected. This is because each point $x_t$ is far from \textit{all} previous points, so necessarily $d(x_t, S_{t-1})^2 > R_t$ is true, resulting in the point being taken. Thus, algorithms using this method (e.g., \citep{liberty2016algorithm, Bhaskara2020}) have center complexities depending on the \textit{aspect ratio}, the ratio between the distances between the furthest two and closest two points in the input sequence.  Data sequences with many scale changes have a provably large aspect ratio, and are unable to be effectively clustered by a direct application of the criteria.

Our important change to overcome the above issue is to use $w_t$ to track the scale. Specifically, during a scale change, we
set $R_t = \frac{w_t^2}{128k}$. By Proposition~\ref{prop:center_vs_mean}, this guarantees that the current value of $R_t$ is at most a factor of $O(n)$ from the optimal value of $\frac{\c_k(X_t)}{k}$. While $O(n)$ seems like a relatively poor approximation, only $O(\log n)$ doublings of $R_t$ are required to increase it to the optimal value, and just $\text{poly}\log n$ centers are taken to do this.

\paragraph{Putting it all together:} Combining our three main ideas, our algorithm consists of the following:

At the start, in lines~\ref{line:init}-~\ref{line:init2}, initialize $S_k, R_k, w_k, \tau_1$ by selecting the first $k$ points and considering the $k+1$th point as the first scale change. Also, initialize the $k$-centers algorithm, which we assume can be given a set of points with the method $online\_k\_centers\_update(x)$ and will return $(Z, w)$: the centers and the cost of the $k$-centers clustering it has computed on all points it has seen so far.

Each time a new point is encountered, update the \textit{scale approximation} (line~\ref{line:k_centers}) and decide if the new point produces a \textit{scale change} (line~\ref{line:begin_if}).

If a scale change is detected, then
\textit{remove centers} that were streamed before the previous scale change and replace them with their corresponding
$k$-centers summary (lines~\ref{line:make_deletion},\ref{line:add_k_centers}).
In the algorithm, we assume that every point streamed has a timestamp that can be accessed with a function denoted as $time$.
We emphasize that when we add $Z_{\tau_{i-1}}$ to $S_t$, the timestamps of the points in $Z_{\tau_{i-1}}$ are from before $\tau_{i-1}$, and thus they will be removed when the next scale change happens.
Furthermore, during a scale change we reset the values of $R_t$ and $F$ (line~\ref{line:updating_old}) to keep these parameters updated for the new scale.

Finally, perform \textit{center selection} using the parameters $R_t,F$ in lines~\ref{line:adding_point}-\ref{line:double_r}.

While the algorithm appears to require us to remember the parameters $R_t, Z_t, w_t, S_t$ for all $t$ and $\tau_{i}$ for all $i$, we may implement it by simply remembering $R_t, w_t, S_t$ at the most recent time, as well as $\tau_{i-1}, \tau_i$, and $Z_{\tau_{i-1}}$ where $i$ is the index of the most recent scale change. This allows us to achieve the desired memory bounds.
\section{Analysis of Algorithm \ref{alg:main}}

The performance of Algorithm \ref{alg:main} is given in Theorem \ref{thm:main}.  Observe that the approximation factor is bounded both in terms of $\c_k(X_n)$, the optimal $k$-centers loss and $\d(X_n)$, our lower bound parameter. The center and memory complexity are both bounded by $O(k \, \text{poly}(\log n))$, and consequently the time complexity of the algorithm is the same. This makes our algorithm the first $O(1)$-approximation in this setting with efficient time and memory complexity. We devote the remainder of our paper to proving Theorem \ref{thm:main}.

Our proof is based on three key steps. First, we bound the online loss $\sum_{t=2}^n d(x_t, S_{t-1})^2$ with three terms: $\c_k(X_n)$, the desired $k$-means loss, $\d(X_n)$, our unavoidable lower bound term, and $\sum_{t=1}^n d(x_t, S_t)^2$. While the last term bears a lot of similarity to our original loss, it has an essential distinction: the term $d(x_t, S_t)^2$ assigns a loss to $x_t$ \textit{after} the algorithm has had a chance to adjust its cluster centers. This is much simpler to analyze, especially in the context of our algorithm as this quantity is precisely $0$ for points that we select. Formally, we have the following proposition (proved in Section \ref{sec:loss_reduce}).
\begin{prop}\label{prop2:lambda_bound}
Suppose we run $\oc(X, k, \delta)$. Then at all times $n$ we have
$$\sum_{t = 2}^n d(x_t, S_{t-1})^2 \leq 8\Lambda(X_n) + 8\sum_{t=1}^n d(x_s, S_t)^2 + 4\c_k(X_n).$$
\end{prop}

Our second step is to bound the loss $d(x_t, S_t)^2$ under the assumption that $R_t$ is sufficiently small for all $t$. This assumption allows us to circumvent the complex way in which the value of $R$ is intertwined with whether or not selections have been made. As a result, we can cleanly divide our analysis into handling the loss, $d(x_t, S_t)^2$ and handling $R_t$ separately. We do so with the following proposition (proved in Appendix \ref{sec:bound_liberty}).

\begin{prop}\label{prop:bound_liberty}
With probability at least $1 -\frac{\delta}{2}$ over the randomness of \oc{}, the following holds simultaneously for all $n \geq 1$: $$\sum_{t = 1}^{n} d(x_t, S_t)^2\ind\left(R_t \leq \frac{\c_k(X_t)}{k}\right) \leq 33\c_k(X_n).$$
\end{prop}

Our third step, is to show that $R_t$ is bounded as indicated in the previous step. We have the following proposition (proved in Appendix \ref{sec:bound_R}).

\begin{prop}\label{prop2:bound_R}
With probability at least $1 - \frac{\delta}{100}$ over the randomness of \oc{}, the following holds simultaneously for all $n \geq k$:
$$R_n \leq \frac{\c_k(X_n)}{k}.$$
\end{prop}

Armed with these three propositions, we have all the ingredients necessary to prove Theorem~\ref{thm:main}. First, a straightforward combination of the three propositions gives us the desired approximation factor of Theorem \ref{thm:main}. Propositions \ref{prop2:bound_R} and \ref{prop:bound_liberty} imply that $\sum_{t=1}^n d(x_t, S_t)^2$ is highly likely to be at most $O(\c_k(X_n))$, and Proposition \ref{prop2:lambda_bound} implies that our online loss consequently satisfies the desired bound.

For the center complexity, memory, and time complexity guarantees of Theorem \ref{thm:main}, we directly derive them from our bound on $R$, Proposition \ref{prop2:bound_R}. The argument is simple: selecting too many points (or equivalently, holding too many points in memory) necessarily increases the value of $R$, which will eventually force the bound in \ref{prop2:bound_R} to be violated. Given that our total number of point selections is small, it also follows that our memory and computation time must be small as well. We now give our proof of Theorem~\ref{thm:main}.

\begin{proof}

We will show that part 1 of Theorem \ref{thm:main} holds with probability at least $1 - \frac{3\delta}{4}$, and parts 2 and 3 of Theorem \ref{thm:main} hold with probability at least $1 - \frac{\delta}{4}$. Theorem \ref{thm:main} will then follow from a union bound.

\paragraph{Approximation Factor:} By a union bound, with probability at least $1 - \frac{3\delta}{4}$, the bounds in Propositions \ref{prop2:bound_R} and Proposition \ref{prop:bound_liberty} both hold. It thus suffices to show that these conditions are sufficient for bounding $\sum_{t=2}^n d(x_t, S_{t-1})^2$. We have,
\begin{equation*}
\begin{split}
\sum_{t=2}^n d(x_t, S_{t-1})^2 &\leq 8\d(X_n) + 8\sum_{t=1}^n d(x_t, S_t)^2 + 4\c_k(X_n) \\
&= 8\d(X_n) + 8\sum_{t=1}^n d(x_t, S_t)^2\ind \left(R_t \leq \frac{\c_k(X_t)}{k}\right) + 4\c_k(X_n) \\
&\leq 8\d(X_n) + 264\c_k(X_n) + 4\c_k(X_n) \\
&= O(\d(X_n) + \c_k(X_n)),
\end{split}
\end{equation*}
where the first inequality holds by Proposition \ref{prop2:lambda_bound}, the second by Proposition \ref{prop2:bound_R}, and the third by Proposition \ref{prop:bound_liberty}.

\paragraph{Center Complexity and Memory:}
For any time $n$, let $U_n$ be those points added to $S$ after time $\tau_i$
and up to time $n$, where $\tau_i \leq n < \tau_{i+1}$.
Namely, $U_n = S_n \cap \{x_{\tau_i}, \ldots, x_n\}$. Because Algorithm~\ref{alg:main}
deletes no points from $S$ between $\tau_i$ and $\tau_{i+1}$, we have
$S_n = (S_{\tau_i}\setminus \{x_{\tau_i}\}) \cup U_n$. At the time of the last scale change $\tau_i$, just before Line~\ref{line:adding_point} is executed,
observe that $(S_{\tau_i}\setminus \{x_{\tau_i}\})
= U_{(\tau_i)-1} \cup \{z_{\tau_{(i-1)}}^1, \ldots, z_{\tau_{(i-1)}}^k\}$. Thus,
$S_n = U_{(\tau_i)-1} \cup U_n \cup \{z_{\tau_{(i-1)}}^1, \ldots, z_{\tau_{(i-1)}}^k\}$.

Now, we will focus on bounding $\max_{n > k} |U_n|$.
Suppose there were a time $n \geq k$ such that $|U_n| \geq 375 k \log^6
(\frac{2n}{\delta})$. Let $\tau_i$
satisfy $\tau_i \leq n \leq \tau_{i+1}$.
By the definition of scale changes and by using Proposition~\ref{prop:center_vs_mean},
we know $256 n w_{\tau_i}^2 \geq w_n^2 \geq \frac{\c_k(X_n)}{n}$.
Therefore, $R_{\tau_i}$, which is set to be $\frac{w_{\tau_i}^2}{128k}$ during the
last scale change of Algorithm~\ref{alg:main}, satisfies $R_{\tau_i} \geq
\frac{\c_k(X_n)}{2^{15}n^2k}$.

Let $f$ be the number of times
that $R$ is doubled from times $\tau_i$ to $n$---because no scale changes occur in this time interval,
we have that $R_n = R_{\tau_i} 2^f$.
Every point in $U_n$ comes from points chosen between times $\tau_i$ and $n$, and
$R$ is doubled at least every $25k \log^5 (\frac{2n}{\delta})$ points that are chosen. Thus,
$f \geq \frac{|U_n|}{25k \log^5 (\frac{2n}{\delta})} \geq 15 \log (\frac{2n}{\delta})$.
However, this implies that $R_n = R_{\tau_i} 2^{15 \log (\frac{2n}{\delta})}
\geq \frac{n^{13}}{\delta^{15}} \frac{\c_k(X_n)}{k} \geq \frac{\c_k(X_n)}{k}$. By
Proposition~\ref{prop2:bound_R},
this event occurs with probability at most $\frac{\delta}{4}$. Thus, the probability
that there exists $n$ such that $|U_n| \geq 375 k \log^6\frac{2n}{\delta}$ is at
at most $\frac{\delta}{4}$.

Thus, $|U_n| \leq O(k \log^6(\frac{n}{\delta}))$ for all $n$ with probability at least
$1 - \frac{\delta}{4}$,
and this implies $|S_n| \leq O(k \log^6(\frac{n}{\delta}))$ with the same probability.
The memory of Algorithm~\ref{alg:main} involves storing $S_n$ and just $O(k)$ additional
points for the $k$ centers and $O(k)$ additional natural numbers bounded by $n$.
Thus, the memory requirement is dominated by $|S_n| = O(k \log^6(\frac{n}{\delta}))$. Finally, it is clear from our algorithm that time time complexity is directly proportional to our memory, which completes the proof.

\end{proof}

\subsection{Proof of Proposition \ref{prop2:lambda_bound}: Bounding the loss with $\d$}\label{sec:loss_reduce}

To prove Proposition~\ref{prop2:lambda_bound}, we introduce a special way of indexing that will make our inequalities more intuitive. Consider the following functions:
\begin{defn}
    Let $X_n$ be a data sequence. For $t \geq 2$, define the previous nearest neighbor
    of the point with index $t$ by $u(t) = \argmin_{i=1,\ldots, t-1} d(x_i, x_t)$; i.e. the index $i$ such that $x_{i} \in X_{t-1}$ is the closest to $x_t$. Consider the tree induced by $u(t)$ where the parent of $t$ is given by $u(t)$.
    Denote the previous sibling in the tree of the point with index $t$ as $p(t) : \mathbb{N} \rightarrow \mathbb{N}$. In other words, $p(t)$ is the greatest index $s$ such that $x_{s} \in X_{t-1}$ satisfies $u(s) = u(t)$---i.e.
    $s$ is the greatest sibling less than $t$. If $t$ has no previous siblings, then set $p(t) = u(t)$.
\end{defn}

\begin{figure}
    \centering
    \captionbox{An illustration depicting $u(t), p(t)$ on a sample dataset.\label{fig:part1}}[0.3\textwidth]{\includegraphics[scale=1]{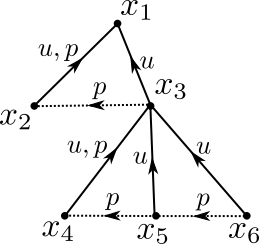}}
    \hspace{0.5cm}
    \captionbox{A naive application of the triangle inequality, which results in
    a $d(x_{u(t)}, S_{t-1})^2$ terms (see~\eqref{eq:cost-reduction}). This produces too many terms involving $x_{u(t)}$.\label{fig:part2}}[0.3\textwidth]{\includegraphics[scale=1]{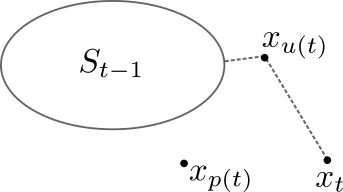}}
    \hspace{0.5cm}
    \captionbox{A double application of the triangle inequality produces
    $d(x_{p(t)}, S_{t-1})^2$ and $d(x_{p(t)}, x_{u(t)})^2$ terms, which involves
    each point a constant number of times (see~\eqref{eq:cost-reduction-2}).\label{fig:part3}}[0.3\textwidth]{\includegraphics[scale=1]{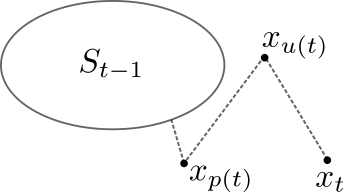}}

\end{figure}

For an illustration of these functions, see Figure~\ref{fig:part1}.
Notice that $\Lambda(X) = \sum_{t = 2}^n d(x_t, x_{u(t)})^2$.
It is easy to see that both $p(t) < t$ and $u(t) < t$ for all $t$. Furthermore, for any index $s$,
there can be at most two distinct indices $t,t'$ such that $p(t) = p(t') = s$, namely,
the very next sibling of $s$ and the smallest child of $s$. We call this property
$2$-injectivity of $p(t)$.

A second important tool we will use is the ability to upper bound $d(x_t, S_{t'})$
in terms of $d(x_t, S_{t})$ for $t \leq t'$.
This follows from the following two cases: if $t,t'$ fall within a scale change, there are no deletions so $S_t$ grows and thus the cost of $x_t$ must shrink. Otherwise, if there are deletions between $t, t'$, we know that old centers in $S_{t}$ are replaced with their $k$-centers approximation in $S_{t'}$, and we can control the induced error accordingly.

\begin{lem}\label{lem:account_for_deletion}
Let $l \geq 1$, and let times $t,t'$ that satisfy
$1 \leq t \leq t' < \tau_{l + 2}$.  Then $d(x_t, S_{t'}) \leq d(x_t, S_t) + w_{\tau_l}\ind(t \leq \tau_{l+1})$.
\end{lem}

\begin{proof}
Suppose $t > \tau_{l+1}$. Then since $t' < \tau_{l+2}$, the latest scale change on or before $t'$ occurs at time $\tau_{l+1}$ during which all points on or before $\tau_l$ are deleted. It follows that $S_t \subseteq S_{t'}$ since no points are deleted between times $t$ and $t'$ inclusive.  This implies the inequality.

Otherwise, if $t \leq \tau_{l+1}$, let $x \in S_t$ be the nearest neighbor of $x_t$ in $S_t$.
If $x$ arrived after $\tau_l$, then $x \in S_{t'}$. Thus $d(x_t, S_{t'}) \leq d(x_t, S_t)$, and we are done. Otherwise,
there is a point $z$ such that $d(z, x) \leq w_{\tau_l}$ is in $S_{t'}$: it is one of the $k$-center points added in Line~\ref{line:add_k_centers} at time $\tau_l$. We can use the triangle inequality to conclude the result.
\end{proof}
Going back to our proof, we need to upper bound $\sum_{t=2}^n d(x_t, S_{t-1})^2$. Let $l$ be such that $\tau_{l+1} \leq n < \tau_{l+2}$.  We begin with a straightforward application of the triangle inequality and Lemma~\ref{lem:account_for_deletion}.
\begin{align}
    d(x_{t}, S_{t-1}) &\leq d(x_{t}, x_{u(t)}) + d(x_{u(t)}, S_{t-1}) \nonumber \\
    &\leq d(x_t, x_{u(t)}) + d(x_{u(t)}, S_{u(t)}) + w_{\tau_l} \ind(u(t) \leq \tau_{l+1}). \label{eq:cost-reduction}
\end{align}
Each $d(x_t, x_{u(t)})$ term is how far $x_t$ is from all other points. Each $d(x_{u(t)}, S_{u(t)})$ term
represents how well $S_{u(t)}$ represents the points $x_{u(t)}$, which is simply the part of $X_{t-1}$ that is closest to $x_t$.
Finally, each $w_{\tau_l} \ind(u(t) \leq \tau_{l+1})$ term represents loss incurred by deleting old points and replacing them with a $k$-centers approximation.
Squaring and summing~\eqref{eq:cost-reduction} over all times $t$, we could obtain a bound for the online loss in terms of how far new points are
from previous points, plus how well $S_t$ represents $X_t$ over time.

However, examining~\eqref{eq:cost-reduction}, a problematic term is $\sum_{t=2}^n d(x_{u(t)}, S_{u(t)})^2$, which
results in a sum of $d(x_u, S_{u})^2$ for all $t$ such that $u(t) = u$. Since $u(t)$
need not be injective, this could produce $n$ copies of $d(x_u, S_u)^2$. To circumvent
this problem, we apply the triangle inequality twice to bound $d(x_{t}, S_{t-1})^2$:
\begin{align}
    d(x_{t}, S_{t-1}) &\leq d(x_{t}, x_{u(t)}) + d(x_{u(t)}, x_{p(t)}) + d(x_{p(t)}, S_{t-1}) \nonumber \\
    &\leq d(x_{t}, x_{u(t)}) + d(x_{u(t)}, x_{p(t)}) + d(x_{p(t)}, S_{p(t)}) + w_{\tau_l}^2\ind(p(t) \leq \tau_{l+1})), \label{eq:cost-reduction-2}
\end{align}
where the first step is the double application of the triangle inequality and the second follows from applying Lemma~\ref{lem:account_for_deletion}. Our first, naive application of the triangle inequality appears in Figure~\ref{fig:part2} and our double application appears in Figure~\ref{fig:part3}. This double application results in terms of the form $d(x_{p(t)}, S_{p(t)})^2$, which when summed over all points, can be upper bounded by the new loss function $\sum_{t = 1}^n d(x_t, S_t)^2$ because $p$ is $2$-injective. Now, we present a full proof.

\begin{proof}
(Proposition \ref{prop2:lambda_bound})
By squaring~\eqref{eq:cost-reduction-2} and applying Cauchy Schwartz, we have $$d(x_t, S_{t-1})^2 \leq 4d(x_t, x_{u(t)})^2 + 4d(x_{u(t)}, x_{p(t)})^2 + 4d(x_{p(t)}, S_{p(t)})^2 + 4w_{\tau_l}^2\ind(p(t) \leq \tau_{l+1}).$$ Substituting this, we form an upper bound on the loss as follows:
\begin{align*}
\sum_{t = 2}^n d(x_{t}, S_{t - 1})^2 &\leq
  \underbrace{4\sum_{t = 2}^n d(x_t, x_{u(t)})^2}_{=4\Lambda(X_n)} +
  \underbrace{4\sum_{t = 2}^n d(x_{u(t)}, x_{p(t)})^2}_{\leq 4\Lambda(X_n)} +
  \underbrace{4\sum_{t = 2}^n d(x_{p(t)}, S_{p(t)})^2}_{\leq 8\sum_{t=1}^n d(x_t, S_t)^2} \\
  &\qquad+ \underbrace{4\sum_{t = 1}^n w_{\tau_l}^2 \ind(p(t) \leq \tau_{l+1})}_{\leq 8\tau_{l+1}w_{\tau_l}^2} \\
\end{align*}
Here, the first equality holds by definition; the second inequality holds
because $\sum_{t=2}^n d(x_{p(t)}, x_{u(t)})^2 \leq \sum_{t=2}^n d(x_{t},
x_{u(t)})^2$ since $p(t)$ maps a child of $u(t)$ to its previous child (and to $u(t)$ if there is no previous child);
the third inequality holds from $2$-injectivity of $p(t)$; and the fourth inequality holds because
$p(t)$ is $2$-injective and maps an index to an index strictly less, and thus at
most $2\tau_{l+1}$ values of $t$ may satisfy $p(t) \leq \tau_{l+1}$.

Finally, by Definition \ref{def2:deletion_times}, $\tau_{l+1}w_{\tau_l}^2 < \frac{w_{\tau_{l+1}}^2}{256},$ and by Proposition \ref{prop:center_vs_mean}, $\frac{w_{\tau_{l+1}}^2}{256} \leq \frac{\c_k(X_n)}{2}.$ Substituting, we obtain $$\sum_{t = 2}^n d(x_t, S_{t-1})^2 \leq 8\Lambda(X_n) + 8\sum_{t=1}^n d(x_t, S_t)^2 + 4\c_k(X_n).$$

\end{proof}

\bibliographystyle{alpha}
\bibliography{references}

\appendix

\section{Proof of Theorem \ref{thm:lower}}\label{app:low_bound}

For convenience, we begin by restating Theorem \ref{thm:lower}.

\begin{repthm}{thm:lower}[\textbf{Lower Bound}]
Let $X$ be any data sequence that contains infinitely many distinct points. Let $A$ be an online clustering algorithm such that its output satisfies $|S_n| \leq b_n$ for all $n$ and for all input sequences, where $\{b_n\}$ is a sequence of positive integers. Then there exists a sequence $\tilde{X} = \tilde{x}_1,\tilde{x_2}, \dots$ such that the following conditions hold.
\begin{enumerate}
	\item $\tilde{X}$ is drawn from the closure of $X$, (i.e. $X$ and its limit points). Thus all points in $\tilde{X}$ are arbitrarily close to points in $X$.
	\item For all $n \geq 2$, the expected loss over $A$ satisfies $\mathbb{E}_A\left[ \sum_{s = 2}^n d(\tilde{x}_s, S_{s-1})^2\right] \geq \Omega(\d(\tilde{X}_n))$.
\end{enumerate}
\end{repthm}

We now prove a specific version of Theorem \ref{thm:lower} for nice sequences (Definition \ref{defn:nice_sequence}).

\begin{lem}[\textbf{Theorem \ref{thm:lower} for nice sequences}]\label{lem:lower_help}
Let $X$ be a nice sequence, and let $A$ be an online clustering algorithm such that its output satisfies $|S_n| \leq b_n$ for all $n$ and for all input sequences, where $\{b_n\}$ is a sequence of positive integers. There exists a sequence $\tilde{X}$ such that
\begin{enumerate}
	\item All elements of $\tilde{X}$ are taken from $X$.
	\item For all $n \geq 2$, the expected loss over $A$ satisfies $\mathbb{E}_A\left[ \sum_{s = 2}^n d(\tilde{x}_s, S_{s-1})^2\right] \geq \frac{\d(\tilde{X}_n)}{24}$.
\end{enumerate}
\end{lem}

\begin{proof}
We generate $\tilde{X}$ recursively with $\tilde{x}_1 = x_1$. Suppose we have generated $\tilde{x}_1, \tilde{x}_2, \dots, \tilde{x}_n$; we will explain how $\tilde{x}_{n+1}$ is obtained.

Let $r_i = \frac{d(x_{i}, x_1)}{4}$ for $i \geq 2$. For $i \neq j$, since $X$ is nice,
\begin{equation*}
\begin{split}
d(x_i, x_j) &> \frac{d(x_i, x_1)/2+ d(x_j, x_1)/2}{2} \\
&= \frac{d(x_i, x_1)}{4} + \frac{d(x_j, x_1)}{4} \\
&= r_i + r_j,
\end{split}
\end{equation*}
which implies $B(x_i, r_i)$ and $B(x_j, r_j)$ are disjoint where $B(x, r)$ denotes the closed ball of radius $r$ centered at $x$. This implies that for \textit{any} set $S_n$ of size $\leq b_n$, there are at most $b_n$ indices $i$ for which $|S_n \cap B(x_i, r_i)| \geq 1$. Thus, for a fixed choice of $S_n$ and a randomly chosen $2 \leq i \leq 3b_n+1$, with probability at least $\frac{2}{3}$, $d(x_i, S_n) > r_i$. Applying this over all $S_n$ generated by $A$ (after seeing $\tilde{x}_1, \ldots, \tilde{x}_n$) and switching orders, we have
\begin{equation*}
\begin{split}
\Pr_i \Pr_A [d(x_i, S_n) > r_i] = \Pr_A \Pr_i [d(x_i, S_n) > r_i] \geq \frac{2}{3}.
\end{split}
\end{equation*}
Thus, there exists $i$ for which $\Pr_A[d(x_i, S_n) > r_i] \geq \frac{2}{3}$. It follows by Markov's inequality that
\begin{equation*}
\begin{split}
\E_A[d(x_i, S_n)^2] &\geq \frac{2}{3}r_i^2  + \frac{1}{3}0 = \frac{d(x_1, x_i)^2}{24}.
\end{split}
\end{equation*}
We now set $\tilde{x}_{n+1} = x_i$, which concludes the definition of $\tilde{X}$. Since $x_1 \in \tilde{X}$, the above equation implies that $$\mathbb{E}_A[d(\tilde{x}_{n+1}, S_n)^2] \geq \frac{d(\tilde{x}_1, \tilde{x}_{n+1})^2}{24} \geq \frac{d(\tilde{x}_{n+1}, X_n)^2}{24}.$$

Summing this and applying linearity of expectation, we have that
\begin{equation*}
\begin{split}
\E_A \left[ \sum_{s = 2}^n d(\tilde{x}_s, S_{s-1})^2 \right] &\geq \sum_{s=2}^n \frac{d(\tilde{x}_{s}, \tilde{X}_{s-1})^2}{24} \\
&\geq \frac{\d(\tilde{X}_n)}{24}.
\end{split}
\end{equation*}
\end{proof}

We now prove Theorem \ref{thm:lower}.

\begin{proof}\textbf{(Theorem \ref{thm:lower})}

We claim there exists a nice sequence of points $Z \subseteq \overline{X}$. By Lemma \ref{lem:lower_help}, this suffices. To show this, we have two cases.

\paragraph{Case 1: $X$ is bounded:} Since $X$ contains infinitely many distinct points, there is an infinitely-long subsequence that has all distinct points. Taking this subsequence if necessary, we assume without loss of generality that all points in $X$ are distinct. Since $X$ is a sequence in a bounded region in $\R^d$, it follows that $X$ is contained within a compact subset of $\R^d$. Thus, by the definition of sequential compactness, $X$ must contain a subsequence that converges. Let $y_1, y_2, \dots $ denote this sequence and let $y$ denote its limit.

In the case that $y = y_i$ for some $i$, simply delete this entry. So we may now assume that $y, y_1, \dots$ are distinct points such that $\lim_{i \to \infty} y_i = y$. We now construct $Z$ recursively. Let $z_1 = y$ and $z_2 = y_1$.

Suppose we have constructed $z_1, \dots, z_n$ thus far. Since $y_i$ converges to $z_1 = y$, we can simply pick a point $y_j$ such that $d(z_1, y_j) < \frac{1}{2}d(z_1, z_n)$. We let $z_{n+1}$ equal such a point, and this concludes the construction.

To verify the condition holds, let $i \neq j > 1$ and without loss of generality suppose $i > j$. This already implies $d(z_j, z_1) > d(z_i, z_1)$, and all that remains is to show that $d(z_i, z_j) > \frac{1}{2}d(z_j, z_1)$. This follows from the triangle inequality:
\begin{equation*}
\begin{split}
d(z_i, z_j) &\geq d(z_j, z_1) - d(z_i, z_1) > d(z_j, z_1) - \frac{1}{2}d(z_j, z_1) \\
&= \frac{d(z_j, z_1)}{2}.
\end{split}
\end{equation*}

\paragraph{Case 2: $X$ is unbounded:} Let $z_1 = x_1$ and $z_2 = x_2$ with $x_2 \neq x_1$. We then construct $Z$ recursively.

Suppose we have constructed $z_1, \dots, z_n$ thus far. Then select $z_{n+1}$ to be any point in $X$ such that $2d(z_1, z_n) < d(z_1, z_{n+1})$. This completes the construction.

To verify the condition holds, let $i \neq j > 1$ and without loss of generality suppose $i > j$. This implies $d(z_i, z_1) > d(z_j, z_1)$, and all that remains is to show that $d(z_i, z_j) > \frac{1}{2} d(z_i, z_1)$. Using the triangle inequality in the same way as in the bounded case, we can establish this. This completes the proof for both the bounded and unbounded cases.
\end{proof}

\section{Proof of Theorem \ref{thm:main}}

\subsection{Proof of Proposition \ref{prop:bound_liberty}: Bounding $\sum_{t=1}^n d(x_t, S_t)^2$}\label{sec:bound_liberty}


\begin{repprop}{prop:bound_liberty}
With probability at least $1 -\frac{\delta}{2}$ over the randomness of \oc{}, the following holds simultaneously for all $n \geq 1$: $$\sum_{t = 1}^{n} d(x_t, S_t)^2\ind\left(R_t \leq \frac{\c_k(X_t)}{k}\right) \leq 33\c_k(X_n).$$
\end{repprop}


Our main idea will be to  fix $n$ (we will later use a union bound to obtain simultaneity over all $n$), and bound the desired sum over subsets of $X_n$ that are between scale changes. This allows us to circumvent issues posed by deleting points. Then, we will sum our bounds over all intervals of scale changes and conclude by appealing to the exponentially growing nature of data over successive scale changes.

We begin with a general lemma that assists in bounding our loss for an arbitrary set of times that occur between of pair of scale changes. Think of $X$ as a fixed, possibly infinite, sequence. Since the online $k$-center subroutine is deterministic, the scale-change times $\tau_i$ are also predetermined. The only randomness arises from line~\ref{line:adding_point} of the algorithm where we probabilistically add a point to $S$.

\begin{lem}\label{lem:liberty}
Let $T \subseteq \{1, \dots, n\}$ be set of times such that $T \subset [\tau_i,
  \tau_{i+1})$ for some $i$. Let $\min T$ denote the smallest time in $T$. Let
   $X_{(T)}$ denote $\{x_t: t \in T\}$. Then for any $\Gamma > 0$ and $\beta \geq 0$, with probability at least $1 - \exp\left(-\frac{\beta \log^4
  \frac{2\min T}{\delta}}{\Gamma}\right)$,
  $$\sum_{t \in T}
d(x_t, S_{t})^2\ind(R_t \leq \Gamma) \leq \beta +\max_{t \in T} \c(X_{(T)}, x_t). $$
\end{lem}

\begin{proof}
  Let $A(T,\Gamma, \beta)$ denote the event that we don't want; i.e. that 
  $\sum_{t \in T} d(x_t,S_{t})^2\ind(R_t \leq \Gamma) > \beta +\max_{t \in T} \c(X_{(T)}, x_t). $
  
  Let $\{E_t\}_{t=1}^\infty$ denote the sequence of Boolean variables indicating
  whether $x_t$ was selected as a center on line~\ref{line:adding_point}. Writing $s = \min T$, we will show the slightly
  stronger statement that $\Pr[A(T, \Gamma, \beta) | E_1, \ldots, E_{s-1}] \leq
  \exp(-\frac{\beta \log^4 \frac{2 s}{\delta}}{\Gamma})$ for any choice of $E_1,
  \ldots, E_{s-1}$. The result will then follow by conditional probability.

We use induction on $|T|$ to show the result holds. For $T = \{\}$, the claim 
  trivially holds. For the inductive step, take $T' = T \setminus \{s\}$; we will suppose that the claim holds for $T'$ and for any choice
  of $\beta', \Gamma'$, and $E_1, \ldots, E_{s'}$, where $s' = \min T'$.
  By marginalizing over the times $s+1$ through $s'$, 
  our inductive hypothesis implies that 
  $\Pr[A(T', \Gamma', \beta')|E_1, \ldots, E_{s}]\leq \exp(-\frac{\beta' \log^4 \frac{2
  s}{\delta}}{\Gamma'})$.

  Given $E_1, \ldots, E_{s-1}$, we have
  \begin{equation}\label{eq:prob_s_chosen}
    \Pr[E_s=0 | E_1, \ldots, E_{s-1}] \leq \max\left\{0, \left(1 - \frac{\gamma^2\log^4
    \frac{2s}{\delta}}{R_s}\right)\right\} \leq \exp\left(- \frac{\gamma^2\log^4
  \frac{2s}{\delta}}{R_s}\right),
  \end{equation}
  where $\gamma = d(x_s, S_s')$, and $S_s' =
  S_s \setminus \{x_s\}$ ($S_s'$ is the value of $S$ before line~\ref{line:adding_point} when deciding to take
  $x_s$).

  We will now bound $\Pr[A(T, \Gamma, \beta) | E_1, \ldots, E_{s-1}]$ as
  follows:
  \begin{align*}
    \Pr[A(T, \Gamma, \beta) | E_1, \ldots, E_{s-1}] &\leq 
    \Pr[E_s = 0 | E_1, \ldots, E_{s-1}]\Pr[A(T, \Gamma, \beta) | E_1, \ldots, E_{s-1}, E_s = 0] \\ 
    &+\Pr[E_s = 1 | E_1, \ldots, E_{s-1}]\Pr[A(T, \Gamma, \beta)| E_1, \ldots, E_{s-1}, E_s = 1].
  \end{align*}
  
  We analyze the two parts separately, beginning with the easier one.
\paragraph{Case 1: $E_s = 1$.} The key observation in this case is that because $T \subseteq [\tau_i, \tau_{i+1})$,  no deletions occur after $x_s$ is selected. Therefore, $x_s \in S_t$ for all $t \in T$ with $t > s$. Furthermore, $x_s$ itself incurs $0$ loss as $x_s \in S_s$. This implies that
\begin{equation*}
\begin{split}
\sum_{t \in T} d(x_t, S_t)^2\ind(R_t \leq \Gamma) &\leq\sum_{t \in T} d(x_t, x_s)^2 \\
&= \c(X_{(T)}, x_s) \\
&\leq \beta + \max_{t \in T} \c(X_{(T)}, x_t).
\end{split}
\end{equation*}
  Thus, $\Pr[A(T, \Gamma, \beta)|E_1, \ldots, E_{s-1}, E_s = 1]
  = 0$.

\paragraph{Case 2: $E_s = 0$.} In this case, we have two subcases, first, when $\gamma^2 > \beta$ and second when $\gamma^2 \leq \beta$.

  First, suppose $\gamma^2 > \beta$. If $R_s > \Gamma$, then $\Pr[A(T,
  \Gamma,\beta)|E_1, \ldots, E_{s-1}, E_s = 0] = 0$ as $A(T, \Gamma, \beta)$
  cannot occur. Otherwise, if $R_s \leq \Gamma$, then using~\eqref{eq:prob_s_chosen}, we have
  \begin{align*}
  \lefteqn{\Pr[E_s = 0 | E_1, \ldots, E_{s-1}]\Pr[A(T,
    \Gamma, \beta) | E_1, \ldots, E_{s-1}, E_s = 0]} \\
    &\leq \Pr[E_s = 0 | E_1, \ldots, E_{s-1}]
    \leq \exp \left(- \frac{\beta\log^4 \frac{2s}{\delta}}{\Gamma}\right).
  \end{align*}

Second, suppose $\gamma^2 \leq \beta$. Recall $T' = T \setminus \{s\}$. Observe that
\begin{equation*}
\begin{split}
  \sum_{t \in T} d(x_t, S_t)^2\ind(R_t \leq \Gamma) &\leq 
  d(x_s, S_s)^2 + \sum_{t \in
  T'} d(x_t, S_t)^2 \ind(R_t \leq \Gamma)\\ &\leq
  \gamma^2 + \sum_{t \in
  T'} d(x_t, S_t)^2 \ind(R_t \leq \Gamma).
\end{split}
\end{equation*}
Since $T' \subset T$, it follows that $\c(X_{(T)}, x_t) \geq \c(X_{(T')}, x_t)$ which implies that $\max_{t \in T}\c(X_{(T)}, x_t) \geq \max_{t \in T'}\c(X_{(T')}, x_t)$.
Combining these observations, we have
\begin{equation*}
\begin{split}
  \sum_{t \in T} d(x_t, S_t)^2 \ind(R_t \leq \Gamma) &\geq \beta + \max_{t \in T}
  \c(X_{(T)}, x_t) \\
  \implies 
  \gamma^2 + \sum_{t \in T'} d(x_t, S_t)^2 \ind(R_t \leq \Gamma) &\geq \beta +
  \max_{t \in T'} \c(X_{(T')}, x_t),
\end{split}
\end{equation*}
so when $\gamma^2 \leq \beta$, $A(T, \Gamma, \beta)$ implies $A(T', \Gamma, \beta-\gamma^2)$.
Thus
\begin{align*}
\lefteqn{\Pr[E_s = 0 | E_1, \ldots, E_{s-1}]\Pr[A(T,\Gamma, \beta) | E_1, \ldots, E_{s-1}, E_s = 0]} \\
&\leq \Pr[E_s = 0 | E_1, \ldots, E_{s-1}] \Pr[A(T', \Gamma, \beta-\gamma^2) | E_1, \ldots, E_{s-1}, E_s = 0] \\
&\leq 
\exp \left(- \frac{\gamma^2\log^4 \frac{2s}{\delta}}{R_s}\right) \exp\left(-\frac{(\beta-\gamma^2) \log^4 \frac{2s'}{\delta}}{\Gamma}\right) \\
&\leq \exp\left(-\frac{\beta \log^4 \frac{2s}{\delta}}{\Gamma}\right) .
\end{align*}
where we have used equation (\ref{eq:prob_s_chosen}) and the inductive hypothesis for the second inequality.
Thus, regardless of whether or not $\gamma^2 \leq \beta$, the bound of
$\exp\left(-\frac{\beta \log^4 \frac{2s}{\delta}}{\Gamma}\right)$ holds, and
we have
\[
  \Pr[A(T, \Gamma, \beta)|E_1, \ldots, E_{s-1}] \leq 
  \exp\left(-\frac{\beta \log^4 \frac{2s}{\delta}}{\Gamma}\right),
\]
as desired.
\end{proof}

Our next step is to apply Lemma \ref{lem:liberty} to get a bound on the loss function over well behaved time intervals. Our key construction for doing this is the notion of a \textit{cluster ring}, which was introduced in \cite{braverman11}.

\begin{defn}\label{defn:cluster_ring}
Let $C$ be a set of points. Let $\mu$ denote the mean of $C$, and $\gamma = \frac{\c(C)}{|C|}$ be the average cost of clustering each point. Then the $j$th \textbf{cluster ring} of $C$, denoted $C_j$, is defined as
\begin{itemize}
	\item $C_0 = \{x \in C :  d(x, \mu)^2 < \gamma\}$
	\item $C_j = \{x \in C: 2^{j-1}\gamma \leq d(x, \mu)^2 < 2^j\gamma\}$ for $j \geq 1$.
\end{itemize}
\end{defn}

The intuition behind cluster rings is that any point in $C_j$ serves as a reasonable cluster center. Thus, cluster rings are particularly amenable to Lemma \ref{lem:liberty}: when $X_{(T)}$ is a cluster ring, the term $\max_{t \in T} \c(X_{(T)}, x_t)$ can be controlled. We apply this in our next step where we consider time intervals $T$ that are both bounded between scale changes and change by at most a factor of two.

\begin{lem}\label{lem:approx_bound_interval}
Let $a, m$ be times satisfying $\tau_i \leq a \leq m < \tau_{i+1}$ for some $i$, and $m < 2a$.
Let $X_{a:m}$ denote $\{x_a, \dots x_m\}$. Then with probability at least $1 - \frac{\delta}{4m^2}$ over the randomness of \oc{},
$$\sum_{t = a}^m d(x_t, S_t)^2\ind\left(R_t \leq \frac{\c_k(X_t)}{k}\right) \leq 8\c_k(X_{a:m}) + 4\frac{\c_k(X_m)}{\log^2 a}.$$
\end{lem}

\begin{proof}
Let $C^1, C^2, \dots, C^k$ denote the optimal $k$-means clustering of $X_{a:m}$, and let $c^1, c^2 \dots c^k$ denote their respective centers. Using Definition \ref{defn:cluster_ring}, let $C_j^i$ denote the $j$th cluster ring of $C^i$. Let $\Gamma = \frac{\c_k(X_m)}{k}$ and $\beta = \frac{3\Gamma}{\log^3a}$. Then by applying Lemma \ref{lem:liberty} to all non-empty $C_j^i$ and applying a union bound, we have that with probability at least $1 - \sum_{i, j: |C_j^i| \geq 1} \exp \left( - \frac{\beta \log^4 \frac{2a}{\delta}}{\Gamma} \right)$
\begin{equation}\label{eqn:using_liberty_bound}
\begin{split}
\sum_{t = a}^m d(x_t, S_t)^2 \ind\left (R_t \leq \frac{\c_k(X_t)}{k} \right) &\leq \sum_{i, j: |C_j^i| \geq 1} \sum_{x_t \in C_j^i}  d(x_t, S_t)^2 \ind\left (R_t \leq \Gamma \right)\\
&\leq \sum_{i, j: |C_j^i| \geq 1} \frac{3\Gamma}{\log^3 a} + \max_{c \in C_j^i}\c(C_j^i, c).
\end{split}
\end{equation}

Note that we can safely replace the indicator variables bounding $R_t$ with a uniform $\ind(R_t \leq \Gamma)$ since $\c_k(X_t)$ is monotonically non-decreasing. 

It consequently suffices to show that $\sum_{i, j: |C_j^i| \geq 1} \exp \left( - \frac{\beta \log^4 \frac{2a}{\delta}}{\Gamma} \right) \leq \frac{\delta}{4m^2}$, and that~\eqref{eqn:using_liberty_bound} implies the desired bound. To do so, we will leverage a few simple properties of cluster rings.

First, observe that there are at most $m$ non-empty cluster rings as there are at most $m$ points in $X_{a:m}$. It follows by substituting this along with $\beta = \frac{3\Gamma}{\log^3 a}$ and $m < 2a$ that 
\begin{equation*}
\begin{split}
\sum_{i, j: |C_j^i| \geq 1} \exp \left( - \frac{\beta \log^4 \frac{2a}{\delta}}{\Gamma} \right) &\leq m\exp \left( - \frac{\beta \log^4 \frac{2a}{\delta}}{\Gamma} \right) \\
&\leq m\exp \left( -\frac{3\Gamma \log^4 \frac{2a}{\delta}}{\Gamma\log^3 a}  \right) \\
&\leq m\exp \left( -3\log \frac{m}{\delta} \right) \\
&\leq \frac{\delta}{4m^2}.
\end{split}
\end{equation*}

Thus,~\eqref{eqn:using_liberty_bound} holds with the desired probability of at least $1 - \frac{\delta}{4m^2}.$ Next, for any $C_j^i$ we upper bound $\max_{c \in C_j^i} \c(C_j^i, c)$. Let $c^i$ denote the optimal cluster center (mean) of $C^i$ and $\gamma_i = \frac{\c(C^i)}{|C^i|}$. Then by Definition \ref{defn:cluster_ring}, we have
\begin{equation}\label{eqn:cluster_bound}
\begin{split}
\max_{c \in C_j^i} \c(C_j^i, c) &= \max_{c \in C_j^i} \sum_{c' \in C_j^i} d(c, c')^2 \\
&\leq \max_{c \in C_j^i} \sum_{c' \in C_j^i} 2d(c, c^i)^2 + 2d(c', c^i)^2 \\
&\leq 4|C_j^i|(2^j\gamma_i) \leq 8|C_j^i|(2^{j-1}\gamma_i) \\
&\leq 8 \sum_{c \in C_j^i} d(c^i, c)^2 = 8\c(C_j^i, c^i). 
\end{split}
\end{equation}

Additionally, there are at most $m - a + 1 \leq a$ points in $X_{a:m}$, which means there are at most $a$ points in $C^i$ for any $i$. It follows that there are at most $\lfloor \log a \rfloor + 1$ non-empty cluster rings, $C_j^i$ as $2^{\log a}\gamma_i$ is too large to be the cost incurred by any point in $C^i$. Applying this along with~\eqref{eqn:cluster_bound}, we have that
\begin{equation*}
\begin{split}
\sum_{i, j: |C_j^i| \geq 1} \frac{3\Gamma}{\log^3 a} + \max_{c \in C_j^i}\c(C_j^i, c) &\leq k(\lfloor\log a \rfloor + 1)\frac{3 \Gamma}{\log^3 a} + 8\sum_{i = 1}^k \sum_{|C_j^i| \geq 1} \c(C_j^i, c^i) \\
&\leq \frac{4\c_k(X_m)}{\log^2 a} + 8\c_k(X_{a:m}).
\end{split}
\end{equation*}
\end{proof}

We are now ready to prove Proposition \ref{prop:bound_liberty}. The main idea is to apply the previous lemma to a series of intervals $[a:m]$ which effectively partition the entire input sequence. While a natural starting point is to simply use the scale changes, $\tau_1, \tau_2, \dots$ (thus considering intervals $[\tau_i:\tau_{i+1}-1]$), this faces a problem; $\tau_{i+1}$ can be potentially much larger than $\tau_i$, and the loss term from Lemma \ref{lem:approx_bound_interval} would have a dependence on $\tau_i$ (as we are implicitly setting $a = \tau_i$). To deal with this, we need to further subdivide the intervals $[\tau_i, \tau_{i+1})$ using a sequence of times
\begin{equation}\label{eqn:subdivision_times}
\tau_i = \tau_{i, 1} < \tau_{i, 2} <  \dots < \tau_{i, s_i+1} = \tau_{i+1},
\end{equation}
that are chosen so that $\tau_{i, j+1} \leq 2\tau_{i,j}$. In the context of Lemma \ref{lem:approx_bound_interval}, this means that $a \simeq m$ up to a constant factor.

\begin{proof}
(Proposition \ref{prop:bound_liberty}). Let $X = \{x_1, \dots\}$ be an input sequence, and let $\tau_1, \tau_2, \dots$ be the scale changes in $X$ (Definition \ref{def2:deletion_times}). As per the discussion above, we begin by defining $\tau_{i,j}$ as follows. 

Let $\tau_{i, 1}, \tau_{i, 2}, \dots, \tau_{i, s_i}$ be defined as $\tau_{i, 1} = \tau_i$, and $\tau_{i, j+1} = \min (2 \tau_{i, j}, \tau_{i+1})$ with $\tau_{i, s_i + 1} = \tau_{i + 1}$ by convention. Note that we define $\tau_l = \infty$ if $\tau_{l-1}$ is the last scale change in $X$, and we correspondingly have that $s_{l-1} = \infty$. 

For any $m \geq k+1$, define $\sigma(m)$ as the largest $\tau_{i, j}$ with $m \geq \tau_{i, j}$. For all $m$, it follows that $\sigma(m) \leq m \leq 2\sigma(m)$ (as otherwise maximality of $\sigma(m)$ would be contradicted) and that no scale changes occur in $[\sigma(m)+1, m]$. Finally, we let $E_m$ denote the event that $$\sum_{t = \sigma(m)}^m d(x_t, S_t)^2 \ind \left(R_t \leq \frac{\c_k(X_t)}{k} \right) \leq 8\c_k(X_{\sigma(m):m}) + 4 \frac{\c_k(X_m)}{\log^2 \sigma(m)}.$$ By Lemma \ref{lem:approx_bound_interval}, $E_m$ holds with probability at least $1 - \frac{\delta}{4m^2}$ which implies (through a union bound) that $\bigcap_{m \geq k+1} E_m$ holds with probability at least $1 - \frac{\delta}{2}$. Thus, it suffices to show that $\bigcap_{m \geq k+1} E_m$ implies that $\sum_{t=1}^n d(x_t, S_t)^2 \ind\left (R_t \leq \frac{\c_k(X_t)}{k}\right) \leq 33 \c_k(X_n)$ holds for all $n$. 

To this end, suppose $\bigcap_{m \geq k+1}E_m$ holds. Fix any $n \geq k+1$ (the case $n \leq k$ is trivial as we pick the first $k$ points by default). Let $\sigma(n) = \tau_{l, r}$. For brevity, we also write $d(x_t, S_t)^2\ind\left(R_t \leq \frac{\c_k(X_t)}{k}\right)$ as $\alpha_t$. It follows that 
\begin{equation}\label{eqn:massive}
\begin{split}
\sum_{t = 1}^n \alpha_t &= \sum_{t=1}^{\tau_l - 1} \alpha_t + \sum_{t = \tau_l}^{n}\alpha_t \\
&= \sum_{t=1}^k d(x_t, S_t)^2 + \sum_{t =\tau_1}^{\tau_l - 1} \alpha_t + \sum_{t = \tau_l}^{n}\alpha_t \\
&= \sum_{i = 1}^{l-1}\sum_{j = 1}^{s_i} \sum_{t = \tau_{i, j}}^{\tau_{i, j+1} -1} \alpha_t +  \sum_{j = 1}^{r-1}\sum_{t = \tau_{l, j}}^{\tau_{l, j+1} - 1} \alpha_t + \sum_{t = \tau_{l, r}}^n \alpha_t. \\
&= \sum_{i = 1}^{l-1}\sum_{j = 1}^{s_i} \sum_{t = \sigma(\tau_{i, j+1} -1)}^{\tau_{i, j+1} -1} \alpha_t +  \sum_{j = 1}^{r-1}\sum_{t = \sigma(\tau_{l, j+1} - 1)}^{\tau_{l, j+1} - 1} \alpha_t + \sum_{t = \sigma(n)}^n \alpha_t,
\end{split}
\end{equation}
with the last step following from $\sigma(\tau_{i, j+1} - 1) = \tau_{i, j}.$ We can now bound the inner summands by noting that each of them corresponds to an event $E_m$. In particular, by applying $E_m$ for $m = \tau_{i, j+1} - 1$ for $1 \leq i \leq l-1$ and $1 \leq j \leq s_i$, we have
\begin{equation*}
\begin{split}
 \sum_{i = 1}^{l-1}\sum_{j = 1}^{s_i} \sum_{t = \sigma(\tau_{i, j+1} -1)}^{\tau_{i, j+1} -1} \alpha_t &\leq  \sum_{i = 1}^{l-1}\sum_{j = 1}^{s_i} 8\c_k\left(X_{\sigma(\tau_{i, j+1} -1):\tau_{i, j+1} -1}\right) + 4 \frac{\c_k\left( X_{\tau_{i, j+1} -1}\right)}{\log^2\sigma(\tau_{i, j+1} -1)}  \\
 &\leq 8\c_k(X_{\tau_l - 1}) + \sum_{i = 1}^{l-1}\sum_{j = 1}^{s_i}4 \frac{\c_k\left( X_{\tau_{i, j+1} -1}\right)}{\log^2\sigma(\tau_{i, j+1} -1)} \\
 &\leq 8\c_k(X_{\tau_l - 1}) + \sum_{i = 1}^{l-1}4\c_k\left( X_{\tau_{i+1}}\right)\sum_{j = 1}^{s_i} \frac{1}{\log^2\tau_{i, j}},
\end{split}
\end{equation*}
with the manipulations coming from combining the $k$-means losses for different intervals of $X$ (i.e. $\c_k(A \cup B) \geq \c_k(A) + \c_k(B)$) and by observing that $\c_k(X_t)$ is monotonic in $t$. 

To further bound this quantity, we simply note that $\tau_{i, j} = 2\tau_{i, j-1}$ for all but the last term, which implies that $\sum_{j = 1}^{s_i} \frac{1}{\log^2 \tau_{i, j}}$ is at most $\frac{7}{4}$ (by crudely bounding the maximal infinite series $\sum \frac{1}{n^2}$). Substituting this gives
\begin{equation*}
\sum_{i = 1}^{l-1}\sum_{j = 1}^{s_i} \sum_{t = \sigma(\tau_{i, j+1} -1)}^{\tau_{i, j+1} -1} \alpha_t \leq 8\c_k(X_{\tau_l - 1}) + 7\sum_{i=1}^{l-1} \c_k\left(X_{\tau_{i+1}}\right)
\end{equation*}
However, the latter sum can be further bounded by observing that by applying Proposition \ref{prop:center_vs_mean} and Definition \ref{def2:deletion_times}, we have
\begin{equation*}
\begin{split}
\c_k(X_{\tau_i}) &\geq \frac{w_{\tau_{i}}^2}{128} > \frac{256\tau_{i-1}w_{\tau_{i-1}}^2}{128} = 2\tau_{i-1}w_{\tau_{i-1}}^2 \geq 2\c_k(X_{\tau_{i-1}}).
\end{split}
\end{equation*}
Thus, by summing a geometric sequence, we have
\begin{equation}\label{eqn:big}
\sum_{i = 1}^{l-1}\sum_{j = 1}^{s_i} \sum_{t = \sigma(\tau_{i, j+1} -1)}^{\tau_{i, j+1} -1} \alpha_t \leq 8\c_k(X_{\tau_l - 1}) + 14\c_k(X_{\tau_l})
\end{equation}
By applying essentially the same argument to the other two sums in Equation \ref{eqn:massive}, we have
\begin{equation}\label{eqn:smaller}
\begin{split}
\sum_{j = 1}^{r-1}\sum_{t = \sigma(\tau_{l, j+1} - 1)}^{\tau_{l, j+1} - 1} \alpha_t &\leq 8\c_k(X_{\tau_l:(\tau_{l, r}-1)}) + 7\c_k\left(X_{\tau_{l,r}} \right) \\
\sum_{t = \sigma(n)}^n \alpha_t &\leq 8\c_k\left(X_{\tau_{l, r}:n}\right) + 4\c_k(X_n).
\end{split}
\end{equation}
Finally, summing Equations \ref{eqn:big} and \ref{eqn:smaller} and combining with Equation \ref{eqn:massive} gives that 
\begin{equation*}
\begin{split}
\sum_{t = 1}^n d(x_t, S_t)^2 \ind\left(R_t \leq \frac{\c_k(X_t)}{k} \right) &\leq 8\c_k(X_n) + 14\c_k(X_{\tau_l}) + 7\c_k(X_{\tau_{l, r}}) + 4\c_k(X_n) \\
&\leq 33\c_k(X_n),
\end{split}
\end{equation*}
completing the proof. 
\end{proof}

\subsection{Proof of Proposition~\ref{prop2:bound_R}}\label{sec:bound_R}

\begin{repprop}{prop2:bound_R}
Running $\oc(X, k, \delta)$ satisfies
$\Pr[R_n \leq \frac{\c_k(X_n)}{k}\text{ for all }n \geq k] \geq 1 - \frac{\delta}{100}$.
\end{repprop}

The proof boils down to showing that once $R_n$ becomes large, \oc{} is unlikely to select many centers after the last scale change. This claim
will be useful later because we will see that selecting lots of centers is the 
main factor increasing the value of $R_n$. Intuitively, this claim is true
because points are selected with probability inversely proportional to $R_n$;
however, proving the claim is complicated by the fact that point
selections are not independent. Thus, we make use of 
martingale concentration results to prove the formal lemma below, though the 
intuition is still straightforward.

\begin{lem}\label{lem:key_lemma}
  Let $n$ be a positive integer such that $n > k$. Suppose we run
  $\oc(X,k,\delta)$, and we are given that there is a time $q < n$ where the
  following hold
  \begin{enumerate}
    \item No scale changes occur in the interval $[q,n]$
    \item $\frac{\c_k(X_n)}{2k} \leq R_t$ for all $t \in [q,n]$.
  \end{enumerate}

  Let $count(q,n)$ be the number of centers selected during the interval
  $[q,n]$. For any $0 < \delta \leq 1$, we have that 
  $\Pr[count(q,n) \geq 25 k \log^5 \frac{2n}{\delta}] \leq
  \frac{\delta}{165 n^2}$.
\end{lem}

\begin{proof}

Let $C^1, C^2, \dots, C^k$ denote the optimal $k$-clustering of $X_n$, and let
  $c^1, c^2 \dots c^k$ denote their respective centers. 
Let $\gamma_i = \frac{\c(C^i)}{|C^i|}$ be the average cost of cluster $C^i$.
For $j \geq 0$, let $C_j^i$ be the $j$th cluster ring of $C^i$ as in
  Definition~\ref{defn:cluster_ring}.
Recall that $C_j^i$ is empty for all $j > \log |C^i|$. Since $|C^i| \leq n$, at most $k(\log n + 2)$ of the sets $C_j^i$ are non-empty.
For ease of notation, for  $x \in X$, let $C(x)$ denote $C_j^i$, where $C_j^i$ is the unique ring with $x \in C_j^i$.

For $q \leq t \leq n$, let $E_t$ be the random
variable defined by $$E_t = \begin{cases} 1 & x_t\text{ is selected, } |C(x_t)
\cap S_t| \geq 1  \\ 0 & \text{otherwise} \end{cases},$$ 
where we take the value of $S_t$ right before Line 16 of the algorithm is executed.

  In other words, if point
  $x_t$ is taken, then $E_t = 1$, except if $x_t$ is the first to be selected in $C(x_t)$. For each ring $C_j^i$, there can be at most one such point.
Since the number of rings is at most $k(\log n+ 2)$, it follows that
$count(q,n) \leq k(\log n+2) + \sum_{t = q}^{n} E_t.$ We will complete the proof by
  showing that $\Pr[\sum_{t=q}^n E_t \geq 24 k \log^5 \frac{2n}{\delta}] \leq
  \frac{\delta}{165 n^2}$.

  We do this by computing an absolute bound on $\Pr[E_t=1]$ for all $q \leq t \leq n$ and then by using a
  martingale concentration result---even though the $E_t$ are not independent,
  the absolute bound still allows us to prove tight concentration.

  Fix $x_t \in C_j^i$. If $x_t$ is the first point in $C_j^i$, then $\Pr[E_t =
  1] = 0$.
  Otherwise, suppose there is a distinct $x_{s} \in C_j^i$ that was selected
  before $x_t$. Because no scale changes occur in the time interval, $x_s \in S_t$ and $d(x_t, S_t) \leq d(x_t, x_s)$.
  By the triangle inequality, $d(x_t, x_s)^2
  \leq 2d(x_t,c^i)^2+2d(c^i,x_s)^2 \leq 2^{j+2}\gamma_i$. This implies that
  \begin{equation}\label{eq:prob_bound}
  Pr[E_t = 1] \leq \frac{d(x_t, S_{t})^2(\log \frac{2t}{\delta})^4}{R_t}
  \leq \frac{2^{j+2}\gamma_i(\log \frac{2n}{\delta})^4}{R_t} \leq
  \frac{8k(2^j\gamma_i)(\log \frac{2n}{\delta})^4}{\c_k(X_n)},
  \end{equation}
  with the last
  inequality holding since $\frac{\c_k(X_n)}{2k} \leq R_{t}$.

  We will apply a standard
  martingale generalization of Bernstein's theorem
  (e.g.~\cite{habib1998methods}, Theorem 3.8), which states that if the random
  variables $E_s$ are zero-one valued, and the maximum possible variance of
  $\sum_{t=q}^n E_t$ is $\hat{v}$, then for all $\lambda > 0$, $\Pr[\sum_{t=q}^n E_t
  \geq \lambda + \E[\sum_{t=q}^n E_t]] \leq \exp(\frac{-\lambda^2/2}{\hat{v} + \lambda/3})$.
  First, we have the following bound on the expectation:

  \begin{equation*}
\begin{split}
  \E[\sum_{t = q}^{n} E_t] &\leq \sum_{i=1}^k \sum_{j=0}^{\log n + 1} \sum_{x_t
  \in C_i^j} \E[E_t] \\
&\leq \frac{8k(\log \frac{2n}{\delta})^4}{\c_k(X_n)} \sum_{i = 1}^k\sum_{j = 0}^{\log n + 1} |C_j^i|2^j\gamma_i \\
&\leq \frac{8k(\log \frac{2n}{\delta})^4}{\c_k(X_n)} \sum_{i = 1}^k\left[\sum_{x \in C_0^i} \gamma_i + \sum_{j = 1}^{\log n + 1} \sum_{x \in C_j^i} 2d(x, c^i)^2\right]\\
&\leq  \frac{8k(\log \frac{2n}{\delta})^4}{\c_k(X_n)} \sum_{i = 1}^k \left[\c(C^i) + 2\c(C^i)\right] \\
&= 24k(\log \frac{2n}{\delta})^4.
\end{split}
\end{equation*}
The main manipulations we make come from the fact that
  $2^{j-1}\gamma_i < d(x, c^i)^2 \leq 2^j\gamma_i$ for all $j \geq 1$, $x \in C_j^i$,
  and from the definition of $\c(C^i)$.

  Next, since each $E_t$ is zero-one valued,
  the variance of any $E_t$ is at most $\sup \Pr[E_t = 1|E_q, \ldots, E_{t-1}]$ which can be upper bounded by~\eqref{eq:prob_bound}. Thus, $\hat{v} \leq \sum_{t=q}^n \sup \Pr[E_t=1 | E_{q}, \ldots, E_{t-1}] \leq 24k \log(\frac{2n}{\delta})^4$ (using the same steps as the expectation
  bound). Applying the martingale form of Bernstein's theorem with
  $\lambda = 24k \log(\frac{2n}{\delta})^4(\log \frac{2n}{\delta}-1)$, we have
\begin{equation*}
\begin{split}
  \Pr\left[\sum_{t = q}^{n} E_t > t + \textstyle{\E\left[\sum_{t=q}^n E_t\right] 
  }\right]
&\leq \exp\left(\frac{-\lambda^2/2}{\hat{v} + \lambda/3}\right) \\
&\leq \exp\left(\frac{-\lambda^2/2}{8k \log^4 \frac{2n}{\delta}(\log \frac{2n}{\delta} + 2)}\right) \\
&\leq \exp\left(-\frac{24^2k (\log \frac{2n}{\delta}-1)^2}{16 (\log \frac{2n}{\delta} + 2)}\right)\\
&\leq \exp\left(-9 \log \frac{2n}{\delta}\right) \\
&\leq \left(\frac{\delta}{2n}\right)^9
\leq \frac{\delta}{165n^2}.
\end{split}
\end{equation*}
This completes the proof.
\end{proof}

To complete the proof of Proposition \ref{prop2:bound_R}, we observe that if $R_n$
becomes larger than $\frac{\c_k(X_n)}{k}$, it could not have been set this high
by Line~\ref{line:updating_old} (as $\frac{w_n^2}{128k}$ is controlled by
Proposition~\ref{prop:center_vs_mean}$)$. Thus, it must be the case that $R_n$
was doubled at time $n$, and that many points were selected since the last scale
change, since the counter $F$ resets in between scale changes. This gives the exact
conditions for Lemma~\ref{lem:key_lemma}, and we are able to bound the probability that $R_n$ becomes large.

\begin{proof}
(Proposition \ref{prop2:bound_R}) 

  Let $A_n$ be the event that $R_n > \frac{\c_k(X_n)}{k}$ and $R_t \leq
  \frac{\c_k(X_t)}{k}$ for all $t < n$---i.e. $n$ is the minimal time for which
  the property $R_n \geq \frac{\c_k(X_n)}{k}$ holds. The events $A_k$ are pairwise disjoint, so we have
  \begin{align*}
    \Pr[\exists n :  R_n > \tfrac{\c_k(X_n)}{k}] = \sum_{n=k}^\infty \Pr[A_n]
  \end{align*}
  Observe that $A_n$ holds only if
  $R_n$ increased via lines~\ref{line:updating_old} or~\ref{line:double_r}. In
  line~\ref{line:updating_old}, it could have been set to
  $\frac{w_n^2}{128k}$, but this quantity is at most $\frac{\c_k(X_n)}{k}$ by
  Proposition~\ref{prop:center_vs_mean}. 

  Thus, $A_n$ holds only if
  line~\ref{line:double_r} is executed at time $n$, so $F_{n-1} + 1 > 25 k \log^5
  \tfrac{2n}{\delta}$. Let $q(n)$ be the largest time less than $n$ for which
  $F_{q(n)} = 0$. The above conditions imply that if $A_n$ holds, then:

  \begin{enumerate}
    \item During the interval $[q(n), n]$, at least $25k \log^5 \tfrac{2n}{\delta}$ 
      centers are taken, because the counter increases every time a center is
      taken. Denote this event by $A^1_n$.
    \item No scale changes occur in $[q(n), n]$, because scale changes reset 
      the counter $F$ to $0$, and thus $q(n)$ is at least the time of the last
      scale change. Denote this event by $A^2_n$.
    \item For all $t \in [q(n),n-1]$, we have $R_{t} = \frac{R_{n}}{2}$ because no
      scale changes occur, and the only time Line~\ref{line:double_r} can
      execute in this interval is at time $n$. This implies
      $\frac{\c_k(X_n)}{2k} \leq R_t$ for all $t \in [q(n),n-1]$. Denote this event
      by $A^3_n$.
  \end{enumerate}

  Conditioning on the value of $q(n)$ above, we have for any $n$,
  \begin{align*}
    \Pr[A_n] &= \sum_{q = k}^n \Pr[A_n | q(n)=q] \Pr[q(n)=q] \\
    &\leq \max_{k \leq q \leq n} \Pr[A_n | q(n)=q] \ \ \ \ \ (\textnormal{because $\Pr[q(n)=\cdot]$
    sum to $1$})\\
    &\leq \max_{k \leq q \leq n} \Pr[A^1_n \wedge A^2_n \wedge A^3_n | q(n)=q] \\
    &\leq \max_{k \leq q \leq n} \Pr[A^1_n | q(n) = q, A^2_n, A^3_n]
  \end{align*}
  However, the inner term $\Pr[A_n^1 | q(n)=q, A_n^2, A_n^3]$ is controlled by
  Lemma~\ref{lem:key_lemma} to be at most $\frac{1}{165n^2}$. Thus, 
  \[
    \sum_{n=k}^\infty \Pr[A_n] \leq \sum_{n=k}^\infty \frac{\delta}{165n^2} \leq
    \frac{\delta}{165}\frac{\pi^2}{6} \leq \frac{\delta}{100}.
  \]
\end{proof}

\end{document}